\documentclass[letterpaper, 10 pt, conference]{ieeeconf}  

\IEEEoverridecommandlockouts                              

\usepackage{graphicx}
\usepackage{epstopdf}
\usepackage{amsmath}
\usepackage{amssymb}
\usepackage{subfigure}
\usepackage{multirow}
\usepackage{pbox}
\usepackage{cite}
\usepackage{algorithm}
\usepackage{booktabs}
\usepackage{wrapfig}
\usepackage{lscape}
\usepackage{algpseudocode}
\usepackage{bm}
\usepackage{url}
\usepackage{esvect}
\usepackage{xcolor}

\DeclareSymbolFont{extraup}{U}{zavm}{m}{n}
\DeclareMathSymbol{\varheart}{\mathalpha}{extraup}{86}
\DeclareMathSymbol{\vardiamond}{\mathalpha}{extraup}{87}

\renewcommand{\baselinestretch}{0.95}

\newtheorem{problem}{Problem}

\DeclareMathOperator*{\argmin}{arg\,min}

\title{\LARGE \bf A Decision Tree-based Monitoring and Recovery Framework for Autonomous Robots with Decision Uncertainties}

\author{Rahul Peddi  and Nicola Bezzo%
\thanks{Rahul Peddi and Nicola Bezzo are with the Departments of Systems and Information Engineering and Electrical and Computer Engineering, University of Virginia, Charlottesville, VA 22904, USA. Email: {\tt \{rp3cy, nb6be\}@virginia.edu}}}

\begin{document}

\maketitle
\thispagestyle{empty}
\pagestyle{empty}
\begin{abstract}
Autonomous mobile robots (AMR) operating in the real world often need to make critical decisions that directly impact their own safety and the safety of their surroundings. Learning-based approaches for decision making have gained popularity in recent years, since decisions can be made very quickly and with reasonable levels of accuracy for many applications. These approaches, however, typically return only one decision, and if the learner is poorly trained or observations are noisy, the decision may be incorrect. This problem is further exacerbated when the robot is making decisions about its own failures, such as faulty actuators or sensors and external disturbances, when a wrong decision can immediately cause damage to the robot. In this paper, we consider this very case study: a robot dealing with such failures must quickly assess uncertainties and make safe decisions. We propose an uncertainty aware learning-based failure detection and recovery approach, in which we leverage Decision Tree theory along with Model Predictive Control to detect and explain which failure is compromising the system, assess uncertainties associated with the failure, and lastly, find and validate corrective controls to recover the system. Our approach is validated with simulations and real experiments on a faulty unmanned ground vehicle (UGV) navigation case study, demonstrating recovery to safety under uncertainties.

\end{abstract}

\section{Introduction} \label{sec:introduction} 


As autonomous mobile robots (AMR) continue to forge their place in our society, we find them performing a variety of tasks that require complex decision-making, such as package delivery, search and rescue, and reconnaissance missions. Many of these robots leverage learning-based decision-making algorithms to make decisions quickly, but most of these algorithms only return one decision, which can often be incorrect due to lack of proper context during training or noise and uncertainty in observations at runtime~\cite{mldecisions}.

In addition, due to the inherent complexity of these systems, a number of factors, such as actuator or sensor faults, can degrade the performance of the robots, which can cause critical damage to the robot itself and compromise its mission. Furthermore, these different failures can often look the same to a human observer and can even cause confusion in state estimation-based~\cite{statest} or bias measurement~\cite{uavdetection} approaches that have been proposed to deal with failures. Learning-based approaches that deal with such problems~\cite{cnnfault,drlfault}, can encode more complex interactions in measurements, and can make better decisions, but even in such critical applications, these only return one decision and do not account for uncertainties. Thus, here we claim that if the robot could assess uncertainties by evaluating other decisions, particularly those that are similar to the initial decision, it might be able to take safer recovery actions.

\begin{figure}[t]
    \includegraphics[width=1\columnwidth]{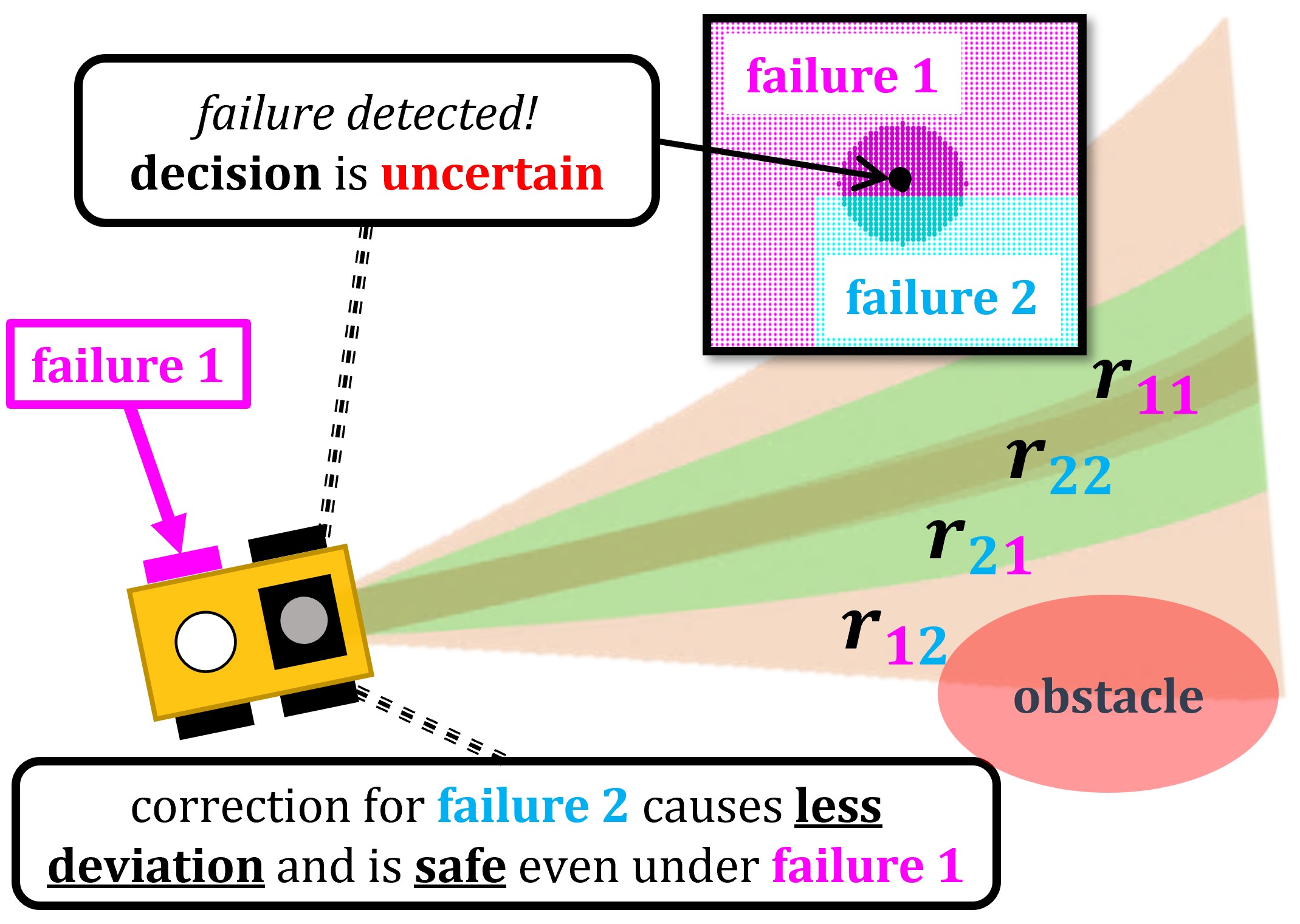}
    \caption{Pictorial representation of the proposed approach: an AMR experiencing either of two failures, evaluates decision uncertainties to find the safest way to correct its behavior, even if the overall performance is temporally degraded.}
    \vspace{-15pt}
    \label{fig:intro}
\end{figure}

In this work, we insist on this principle and investigate the case in which a robot must detect a system failure (e.g., on its sensors and actuators), if one is present, and account for decision uncertainties to safely correct its behavior. Specifically, we propose a novel uncertainty-aware and explainable decision tree (DT)-based monitor to detect at runtime which failure is affecting the system, and what other failures may be plausible given the uncertainties in the initial DT detection. Reachability analysis (RA) is then leveraged to identify safe corrective measures within a library of pre-trained Model Predictive Controllers (MPCs), which are selected in this work because of their capability of predicting future states. Finally, a Bayesian performance validation scheme is proposed to reinforce (or decrease) confidence in the selected corrective measure. As a complementary effect, differently from other learning-enabled methods, a human-interpretable explanation is generated which can potentially be leveraged by a human operator to further improve uncertainty assessment and validation.
 
Shown in Fig.~\ref{fig:intro}, is a pictorial demonstration of our approach, in which an AMR detects failure $1$ and identifies that the detection is very close to a decision boundary between failure 1 and 2. Then the AMR computes the reachable sets for different controller-failure combinations, and chooses a corrective measure that is tuned for failure $2$ even though the system is undergoing failure $1$. By applying such correction, the reachable sets $r_{21}$ (related to the correction 2 on the robot with failure 1) and $r_{22}$ (related to the correction 2 on the robot with failure 2) keeps the system safe even in case of incorrect detection. In contrast, if correction 1 was chosen while failure 2 was occurring, the robot could have collided with the obstacle as depicted by the reachable set $r_{12}$.




This work presents three main contributions: 1) the design of an explainable DT-based monitor that detects if the system is experiencing a failure and the type of failure, 
2) a perturbation-based method to assess uncertainties in decision-making with a reachability-based method to find safe corrective measures given decision uncertainties and 3) a Bayesian validation scheme to increase/decrease confidence in the selected corrective measure.


\section{Related Work} \label{sec:relatedwork}

Decision making for AMR has become a well-studied problem over the years~\cite{esenram}, but safe decision making under uncertainties remains an open challenge. Many recent approaches use learning enabled components, such as deep neural networks (DNN)~\cite{cnnfault} and deep reinforcement learning (DRL)~\cite{drlfault}, to make quick decisions with a reasonable level of accuracy for many applications~\cite{mldecisions}. However, a vast majority of these techniques do not consider uncertainties and return only one decision, which might be incorrect in the presence of measurement or process noise at runtime~\cite{incorrectpred}. Significant effort has been devoted to achieving uncertainty-aware decision making with machine learning; authors in~\cite{drl_uncert,uncertaintyquant} use sampling-based methods. However, the effectiveness of sampling-based methods for uncertainty evaluation relies on the quality and number of samples taken and can become too computationally expensive for robot control in many cases~\cite{samp_slow}. More recently, Bayesian Neural Networks (BNN)~\cite{bnn1} have also been proposed to effectively identify decision uncertainties, however, BNNs are difficult to implement and train due to their complexity, and much of the current research involves finding techniques to make BNNs easier to train~\cite{bnntut}. In contrast to BNNs, we leverage decision trees (DT), which are much simpler to implement and train, and we exploit decision boundaries within the DT framework to assess uncertainties.

Additionally, the aforementioned methods using DNNs, DRL, and BNN contain black boxes, which make it difficult for a user to understand why a particular decision was made, which has shown to improve the overall performance of decision-making systems~\cite{whyblackbox}. Other approaches make considerations on the training dataset through variational inference~\cite{varinfo} on the training data and active learning~\cite{activelearn} to perturb and gain more information about decisions. We take inspiration from these approaches and integrate them into our explainable approach for failure detection and recovery.

As for detecting and recovering from sensor and actuator failures~\cite{jintfault}, control theorists have proposed a number of approaches, in which detection often relies on state estimation~\cite{statest} and deviation/bias measurement and analysis~\cite{uavdetection}, which are easy to understand and work well for detecting different degrees of particular types of failure, but do not extend well to detecting different failures that can appear similar, thus making learning-based approaches more appealing for our case study~\cite{rnnfault}. The control techniques used for correction include adaptive control~\cite{ccdc2019} and model predictive control (MPC), which has been shown extensively to produce safe motion planning under degraded conditions~\cite{mpcsurvey}. In this work, we bridge the gap between learning and control based approaches by designing an explainable Decision Tree (DT)-based monitor for learning-based decision-making without the use of black boxes and integrating MPCs designed to keep the system safe under different sensor and actuator faults and disturbances.

\section{Problem Formulation} \label{sec:probform}
Consider an autonomous mobile robot (AMR) that is navigating to a goal location. The dynamics of the system under nominal conditions can be represented in the state space form, $\bm{\dot{x}} = g(\bm{x}, \bm{u})$ where $\bm{x}$ is the state and $\bm{u}$ is the input, which is set by the control law $\bm{u} = c_0(\bm{x})$. The system, however, can experience a number of different failures, including faulty sensors or actuators (wheel encoders, propellers) or environmental disturbances (wind, ice) that cause the system to deviate from its nominal behaviors. One of these failures may change the dynamics to $\bm{\dot{x}} = g'(\bm{x}, \bm{u})$, and since the control law $\bm{u} = c(\bm{x})$ was tuned for the nominal dynamics, can lead to unsafe behaviors if used to determine the input for the system under a failure. In this work, we assume a set of predefined control laws for different failures is designed, and the main challenge we focus on is deciding which one to use given an unknown failure. This is largely because distinguishing between different types of failures can be very challenging for a human observer, standard control-based failure detection, and even some learning-based failure detection methods, if poorly trained, because of similarities between different failures, and in many cases, waiting until differences appear may not be safe. We also note that in many cases, failures have been experienced before and corrective measures that maximize performance and safety for each failure can be prepared proactively. The challenge then becomes finding a technique to detect which of these failures most closely represents the failure that might be affecting the AMR, and if the uncertainty is high, assess which of the predefined control laws to use to keep the system safe while collecting more data to make a more informed decision to safely recover the degraded system.
\textbf{{\problem{Uncertainty Aware Failure Detection and Recovery:}}} Consider an autonomous robot tasked to navigate through a cluttered environment under the effect of an unknown failure $f_i$ which can degrade its motion performance.  Consider a set of sensor or actuator failures $\mathcal{F}=\{f_1, f_2,...f_N\}$ with associated corrective measures in the form of control laws $\mathcal{C}=\{c_1, c_2,...c_N\}$. The objective of this work is to design a framework to detect the set of possible failures $\bm{f}\subseteq\mathcal{F}$ that explain the behavior of the robot undergoing failure $f_i$ and determine the appropriate control policy $c^* \in \mathcal{C}$ that minimizes tracking error and maximizes safety (i.e., avoids collision with surrounding obstacles):
\begin{equation}
    ||\bm{x}(t)-\bm{x}_r(t)|| = 0, \text{~as~} t\to\infty
\end{equation}
\begin{equation}
    ||\bm{x}(t)-\bm{o}_i|| > 0, ~\forall i=[1,\ldots,N_o]
\end{equation}
where $\bm{x}(t) = [x,y]^{\top}$ is the position of the robot at time $t$, $\bm{x}_r(t)$ is the desired reference state of the robot, and $\bm{o}_i(t)$ is the position of the $i^{th}$ obstacle and $N_o$ is the number of obstacles.
\section{Approach} \label{sec:approach}

\begin{figure}[h]
    \centering
    \includegraphics[width=0.48\textwidth]{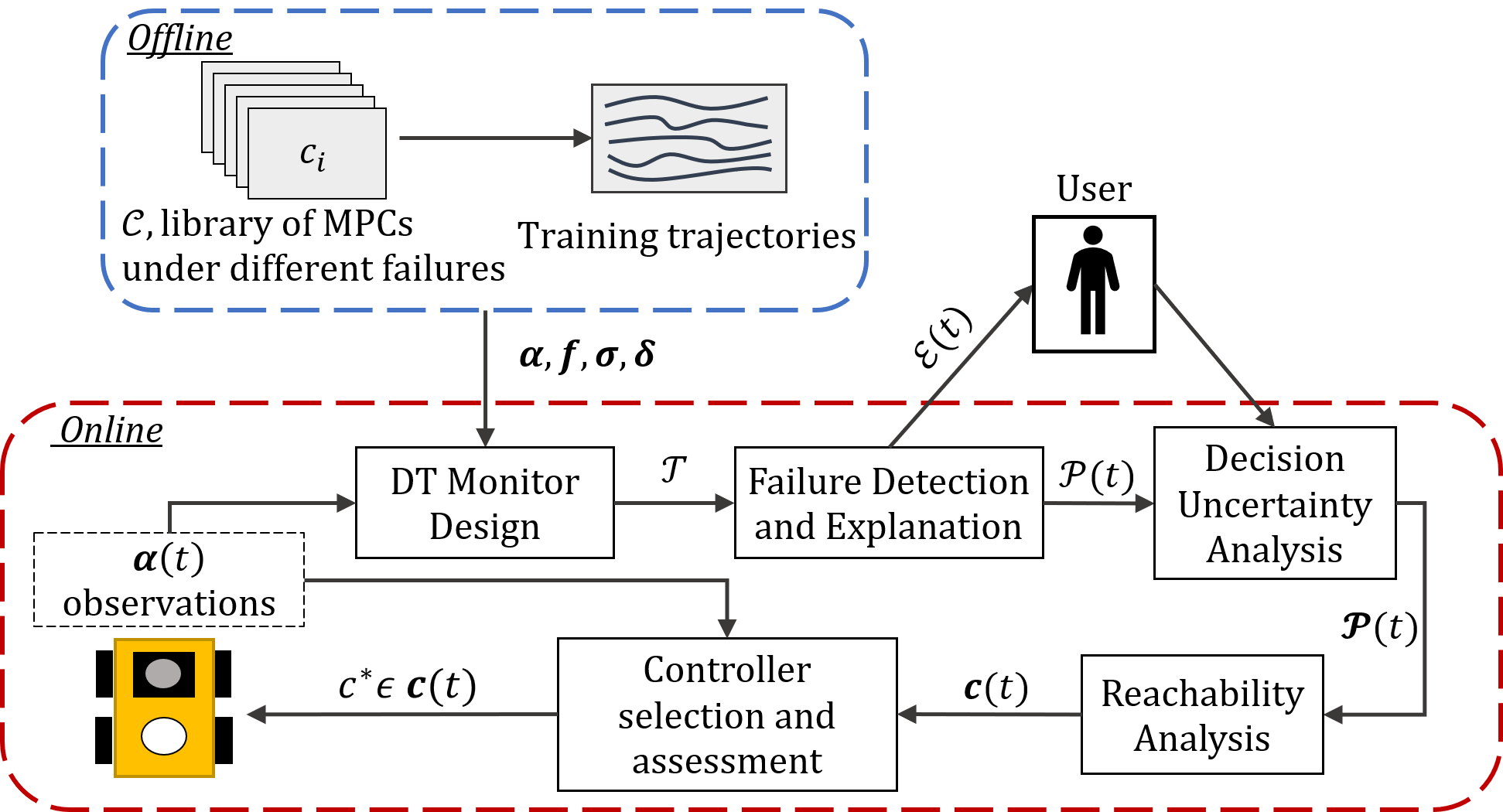}
    \caption{Block diagram of proposed approach.}
    \label{fig:bd}
\end{figure}

In this section, we describe our framework for safe recovery of AMR navigation operations under degraded conditions caused by faulty actuators/sensors or environmental conditions. Our framework consists of offline and online stages, as demonstrated in Fig.~\ref{fig:bd}.

In the offline stage, a robot performs navigation tasks using a model predictive controller (MPC) due to its model-based predictive properties, that can be leveraged to detect failures at runtime. The robot is faced with different actuator and sensor failures, and a different MPC is tuned and tested to maintain a desired level of performance under each failure. This set of MPCs, $\mathcal{C}$, is used to generate the training trajectories, in which we collect observations, $\bm{\alpha}$, associated failures, $\bm{f}\in\mathcal{F}$, deviations observed for each controller-failure combination, $\sigma_{cf}\in\bm{\sigma}$, and a local perturbation distance for each observation $\bm{\delta}$, which is used at runtime to assess uncertainty in decision-making.

We build a decision tree (DT)-based monitor, $\mathcal{T}$, to detect at runtime which failure, $\mathcal{P}(t)\in\mathcal{F}$, might be affecting the system, and to compute an explanation, $\mathcal{E}(t)$ for this decision, which is communicated to a human user for verification. The monitor output and user input are then used for uncertainty analysis, which determines an additional set of failures, $\bm{\mathcal{P}}(t)\subseteq\mathcal{F}$, that might be possible. Reachability analysis is then used to identify a set of safe corrective measures $\bm{c}(t)\subseteq\mathcal{C}$. Each corrective measure is then assessed for confidence and runtime deviations to select one that is most appropriate for the detected failure and uncertainties. This procedure is repeated, constantly re-evaluating predictions and explanations to gain confidence in the robot's decision-making and converge to safe robot behaviors under degraded conditions. In the next sections, we describe in detail each part of our approach.

\subsection{Baseline Model Predictive Controller} \label{sec:mpcplan}


A set of model predictive controllers (MPC), $\mathcal{C}$ is designed to deal with the different failures we consider in this work. Each controller $c_i\in\mathcal{C}$ is tuned appropriately based on the dynamics of each degraded system. In this work, we use standard MPC for trajectory tracking~\cite{mpcsurvey} since it inherently provides predictions for the robot's future states $\bm{x}_p(t) = [x~y~\theta]^{\top}$, which will be compared with the observed state $\bm{x}(t)$ of the robot at runtime to facilitate failure detection and will be used for the reachability analysis performed in Sec.~\ref{sec:reach}.

\begin{figure}[h]
    \centering
    \includegraphics[width=0.48\textwidth]{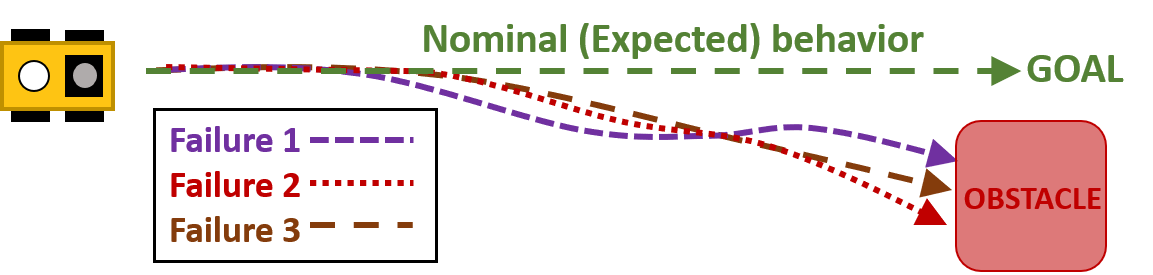}
    \caption{Examples of robot behaviors under different failures, showing intertwining trajectories and deviations with different colliding behaviors.}
    \label{fig:diffault}
\end{figure}

\begin{figure}[h]
    \centering
    \includegraphics[width=0.48\textwidth]{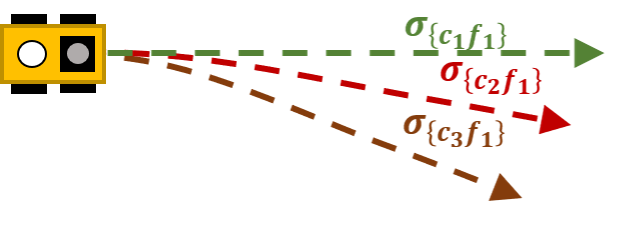}
    \caption{Examples of deviations obtained with different controllers on a particular failure.}
         \vspace{-15pt}
    \label{fig:devex}
\end{figure}

A key focus in this work is recovering failures that can appear very similar to a human observer, as demonstrated by the intertwined deviating trajectories in Fig.~\ref{fig:diffault}.
Robots undergoing each of these failures may have different dynamic models, or may require different weighting parameters to achieve accurate reference tracking, and applying the incorrect controller to a misinterpreted failure may result in unsafe conditions. Thus training is performed with all combinations of failures and controllers to assess what deviations $\bm{\sigma}_{cf}$ may appear if an incorrect control policy is applied to a particular failure. Shown in Fig.~\ref{fig:devex} is a pictorial example of deviations that are observed when testing several controllers on a particular failure.

\subsection{Decision Tree Detections and Uncertainty Assessment}
To detect which failure is affecting the system at runtime, we design an interpretable monitor that leverages decision trees (DT), which are a form of supervised learning that consist of interpretable white-box models~\cite{rahulral}. DTs take in a set of input variables to make a prediction about some output. DTs are made up of a network of nodes, and the outermost nodes, known as leaves, correspond to labels given in the training (failures, in this work). In our failure detection case study, the input variables were found through experimental evaluation and are defined as follows:
\begin{equation} \label{eq:attributes}
\bm{\alpha} = [\Delta x~~\Delta y~~ \Delta \theta~~ c_i]
\end{equation}
where $\Delta x$, $\Delta y$, $\Delta \theta$ are deviations between the predicted state, $\bm{x}_p(t)$ of the MPC and the observed state, $\bm{x}(t)$ of the robot, and $c_i\in\mathcal{C}$ is the controller being deployed by the robot at the time of detection. The controller $c_i$ is included as a categorical predictor~\cite{catvar}, which is discrete and serves to better detect failures when any of the controllers are being used, since different deviations can be expected when different controllers are deployed under each failure. 

\begin{figure}[h]
    \centering
    \includegraphics[width=0.48\textwidth]{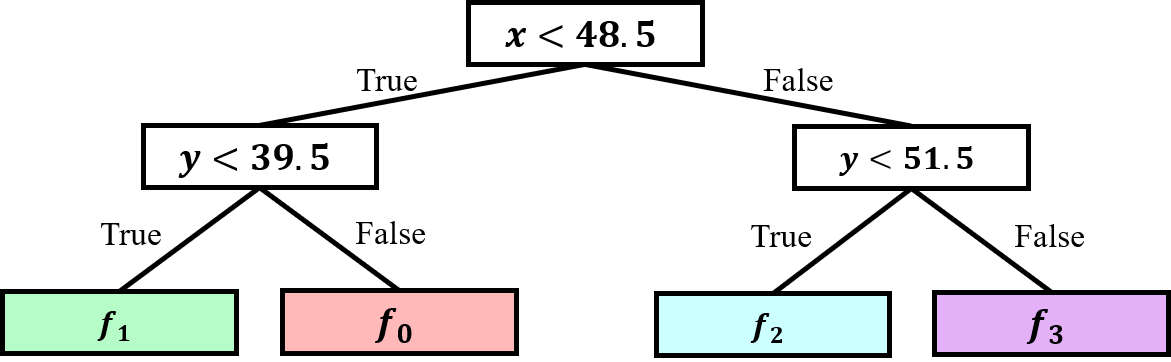}
    \caption{DT used for initial failure detection.}
    \label{fig:tree}
     \vspace{-10pt}
\end{figure}

The training process consists of testing each controller $c_i\in\mathcal{C}$ on an AMR undergoing each failure in $\mathcal{F}$ in both simulation and in hardware experiments, since testing and results are shown in both domains. The outcome of the training consists of the attributes $\bm{\alpha}$ collected at each iteration and each associated ground truth label $f_i$. Each pair of attributes and labels will be denoted as a sample $s_i$, and the collection of all samples (i.e., the entire training set) is denoted as $\mathcal{S}$. A decision tree, $\mathcal{T}$, is grown using the training data, and after taking an observation, $\bm{\alpha}(t)$, an initial failure detection can be obtained:
\begin{equation} \label{eq:pred}
    \mathcal{P}(t) = \mathcal{T}(\bm{\alpha}(t)) = f_i \in \mathcal{F}
\end{equation}
After making the initial decision, a human readable explanation $\mathcal{E}(t)$ for this decision is computed by traversing the path $\Gamma$ from the root of the tree, $\mathcal{V}_0$, to a prediction leaf, $\mathcal{V}_p$, taking the conjunction of each split condition, $c$, for the $N_i$ nodes along the path:
\begin{equation}\label{eq:explanation}
    \mathcal{E}(t) = \bigwedge_{k=1}^{N_i} {c}_k \quad \textrm{with}\quad \Gamma ~| ~\mathcal{P}(t)
\end{equation}
Shown in Fig.~\ref{fig:tree} is a simple example of a DT used to make an initial detection with attributes $[x~y]=[48~40]$. The failure detected is $\mathcal{P}(t) = f_0$, and through \eqref{eq:explanation}, the following explanation is obtained:
$\mathcal{E}(t)=f_0$ because: $\{x<48.5 \land y>39.5 \}$.
The decision obtained from this procedure, however, does not account for uncertainties, and as a result, provides information about only one outcome of the DT, which can be incorrect in the presence of noise and uncertainty. Thus, it is critical to understand what additional failures might be present by assessing these uncertainties.

\subsubsection{Perturbation Based Uncertainty Assessment}
\begin{figure}[h]
	\subfigure[$\delta = 8$ \label{fig:2dbound}]{\includegraphics[width=0.49\columnwidth]{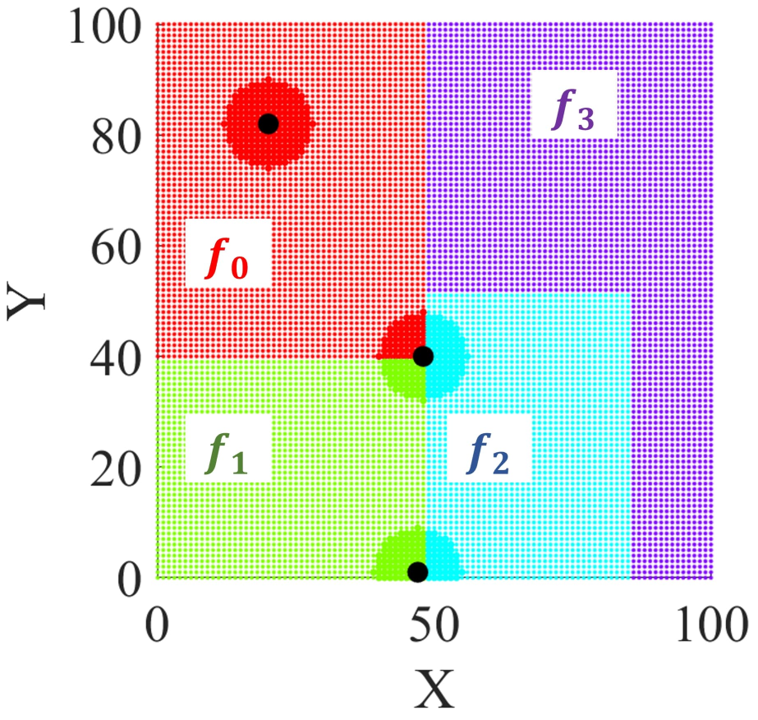}}
	\subfigure[$\delta = 16$ \label{fig:3dbound}]{\includegraphics[width=0.49\columnwidth]{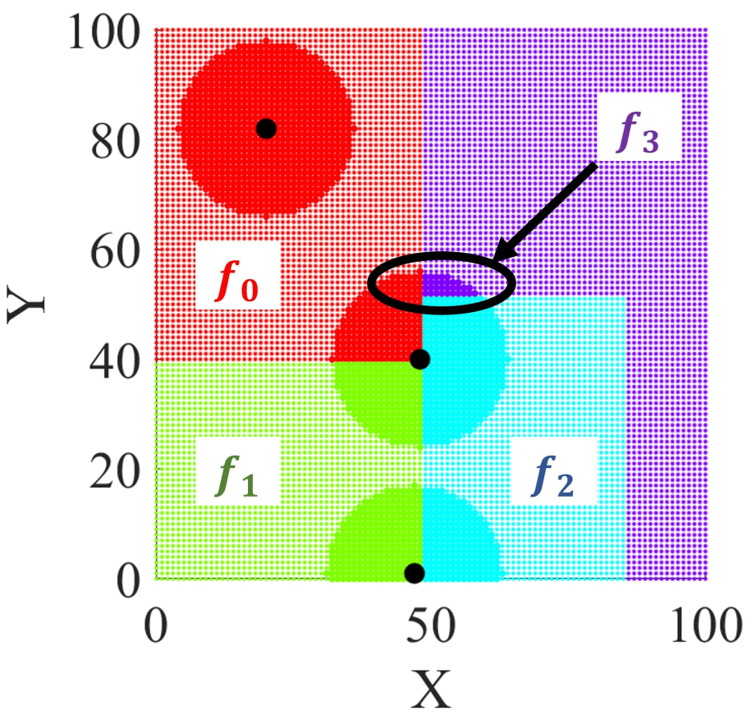}}
	\caption{Examples of local perturbations of different $\delta$ for multiple data points (black points).}
	\label{fig:dbounds}
\end{figure}

To facilitate uncertainty assessment at runtime, uncertainties are first quantified in the training by computing a local perturbation distance $\delta_{s_i}$ that characterizes the distribution of the dataset in the region around each training point $s_i$. In general, dense regions contain more context, resulting in accurate decisions with more certainty and vice versa. However, distance to decision boundaries plays an important role in determining uncertainty; even a decision taken in a dense region close to decision boundaries may be incorrect due to noise. To capture these uncertainties, $\delta_{s_i}$ is defined as the radius of the smallest region around training data point $s_i$ that contains $N_s$ observations, where $N_s$ is a user-defined parameter that depends on the overall quality of the training data and the available computational resources~\cite{dtrees}.
The local perturbation distance, $\delta^*$, for the runtime observation, $\bm{\alpha}(t)$, is computed by finding the corresponding value of the closest training point in $\mathcal{S}$:
\begin{equation}
    \delta^* = \delta_{s_i},~\text{where}~s_i = \argmin_{s_i}||\bm{\alpha}(t)-s_i|| ~\forall s_i \in \mathcal{S}
\end{equation}
Using the computed perturbation distance, which characterizes uncertainties around $\bm{\alpha}(t)$, a perturbed dataset containing input observations and outputs (different failures), is found as follows:
\begin{equation} \label{eq:local_selection}
\bm{s}(t)\subset\mathcal{S} \;\textrm{s.t.}\, ||\bm{\alpha}(t)-s_i|| \leq \delta^* ~\forall s_i \in \mathcal{S}
\end{equation}
Collected from within the perturbed dataset is $\bm{\mathcal{P}}(t)$, which we define as the set of all possible failures for a given set of attributes and associated uncertainties. Shown in Fig.~\ref{fig:dbounds} is an example of a uniformly distributed training dataset with two attributes, $\bm{\alpha} = [x~y]$ and four outputs, indicated by the colored regions, akin to the DT in Fig.~\ref{fig:tree}.
The highlighted smaller circles inside each figure show local perturbations with different $\delta$ around the black observations. It is evident that with larger perturbations (right), another output ($f_3$) is included in the local region for the data at $[x~y] = [48~40]$, due to the proximity to the decision boundaries, indicating that $f_3$ should be included in $\bm{\mathcal{P}}$ given the observation and $\delta=16$. It should be noted that we show this simple toy example to help illustrate our perturbation in a more legible way, but in the implementation, the perturbation radius considers more attributes~\eqref{eq:attributes} that define decision boundaries, making visualizations more cluttered.

Also included in this work is the ability for a human user to intervene based on the original failure detection, $\mathcal{P}(t)$ ,and explanation, $\mathcal{E}(t)$, to modify $\bm{\mathcal{P}}(t)$. For example, if the user decides that the system should be certain about the initial detection, then they can set $\bm{\mathcal{P}}(t) \xrightarrow{} \mathcal{P}(t)$. The user can also add failures to $\bm{\mathcal{P}}(t)$ if their judgement suggests it is needed. In this way, the proposed approach is able to further make use of the readability and interpretability of the DT monitor towards recovering the degraded AMR.

\subsection{Reachability Analysis and Controller Selection} \label{sec:reach}

\begin{figure}[h]
    \centering
    \includegraphics[width=0.44\textwidth]{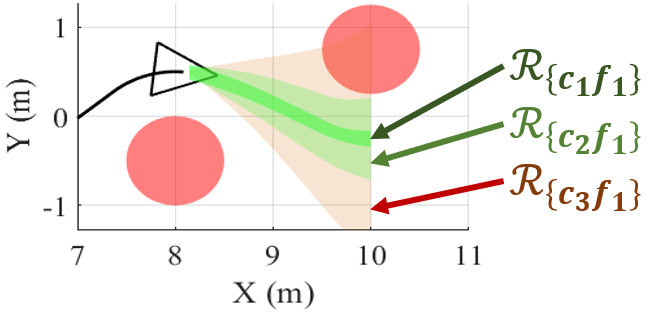}
    \caption{Examples of reachable sets obtained with different controllers on a particular failure.}
    \label{fig:reachex}
\end{figure}

To recover the AMR, we first use reachability analysis (RA)~\cite{reach} to identify a set of controllers $\bm{c}(t)\subseteq\mathcal{C}$ that can be safely deployed for any of the possible failures. Reachable sets $\mathcal{R}_{cf}$ for all combinations of controllers and failures in $\bm{\mathcal{P}}(t)$ are computed by interpolating a region around the MPC predictions $\bm{x}_p({k})$ with $~k = [t,t+N]$, where $N$ is the MPC prediction horizon. The reachable set is bounded by maximum deviations $\bm{\sigma}_{cf}$ collected in the training. Then, the set of safe controllers, $\bm{c}(t)$ is determined by verifying that each reachable set is within an obstacle free region $\mathcal{X}(t)$:
\begin{equation}
    \bm{c}(t) = c_i| \mathcal{R}_{c_if_j}(t) \subseteq \mathcal{X}(t),~\forall c_i|f_i \in \bm{\mathcal{P}}(t),~\forall f_j \in \bm{\mathcal{P}}(t)
\end{equation}
Fig.~\ref{fig:reachex} displays an example taken from our simulations, in which each reachable set is obtained with different controllers assuming failure 1, and since $\mathcal{R}_{c_3f_1}$ intersects with an obstacle, it is unsafe and $c_3\not\in\bm{c}(t)$.

Each controller within $\bm{c}(t)$ can be safely tested, and given no other context, the controller selected, $c^*\in\bm{c}(t)$, is a conservative one with the lowest worst-case deviation for all failures in $\bm{\mathcal{P}}(t)$.

\subsubsection{Controller Confidence Assessment and Validation} \label{sec:runtime}
Selecting a corrective measure as described above is conservative and may not necessarily maximize the performance of the degraded AMR. To improve its performance, the robot needs to gain context at runtime about the effectiveness of the selected controller and decide whether to switch to another. We model this context as a controller confidence $\mathrm{Pr}(c_i)$ that evolves over time based on the observed runtime deviation, which is computed as follows:
\begin{equation}
    \eta_{c_i}(t) = ||\Delta x ~ \Delta y ~ \Delta\theta||
\end{equation}
If using controller $c_i$ causes less deviation than a pre-defined desired level of performance, $\eta_{c_i} < \eta^*$, the confidence in $c_i$ should be reinforced, and vice versa. This controller confidence is formulated as an unknown discrete probability mass function (PMF) over the different controllers with uniform initial confidence estimates: $\mathrm{Pr}(c_i) = 1/N_c,~\forall c_i \in \mathcal{C}$, where $N_c$ is the total number of controllers in $\mathcal{C}$. To reflect the desired effect of growing confidence with positive reinforcement and vice versa, $\mathrm{Pr}(c_i)$ is updated using recursive Bayesian inference~\cite{bayes}, given as follows:
\begin{equation} \label{eq:bayup}
    \mathrm{Pr}(c_i|\rho_{c_i}) = \dfrac{\mathrm{Pr}(\rho_{c_i}|c_i)\mathrm{Pr}(c_i)}{\beta}
\end{equation}

where $\rho$ represents whether the performance criterion $\eta_{c_i}< \eta^*$ is met, and $\beta$ is a normalization constant that is used to ensure that the integral of the discrete PMF is $1$. A confidence threshold, $\gamma$, is set to determine and select the controller that performs best for a given failure, $c^* = ~c_i | \mathrm{Pr}(c_i) \geq \gamma$.
In this work, we set $\gamma = 0.75$, but this choice depends on the application; higher values will be more conservative and vice versa. When a controller meets this threshold, it is automatically deployed given that it is in $\bm{c}(t)$, regardless of the output of the DT monitor. When a controller is deployed and reinforced negatively, it is penalized according to~\eqref{eq:bayup}, and if all controllers cause deviations $\eta_{c_i}>\eta^*, ~\forall c_i\in\bm{c}(t)$, then confidence for all controllers is reinitialized in order to allow the system to test other controllers again based on which has caused the lowest deviations at runtime: $c^* = \argmin_c\bm{\eta}$, where $\bm{\eta}$ is the set of most recently observed deviations for each controller in $\bm{c}(t)$.
This type of situation arises when an unknown failure appears at runtime and every controller creates deviations $\eta_{c_i}>\eta^* ~\forall c_i\in\bm{c}(t)$, but runtime deviations and reachable sets suggest the system can still remain safe. We show one such case in Section~\ref{sec:results}.1. In a situation where no safe controller exists, $\bm{c}(t) = \varnothing$ for all failures in $\bm{\mathcal{P}}(t)$, the robot should switch into an ad-hoc fail safe mode, which can vary based on the application.
\section{Results} \label{sec:results}

The case study investigated in this work and presented in this section consists of an AMR navigation task in the presence of different sensor and actuator failures. The robot is expected to detect the failure affecting the system, assess uncertainties around this initial decision, and recover to a safe mode of operation.

\subsubsection{Simulations} \label{sec:sims}
\begin{figure}[h]
    \includegraphics[width=0.48\textwidth]{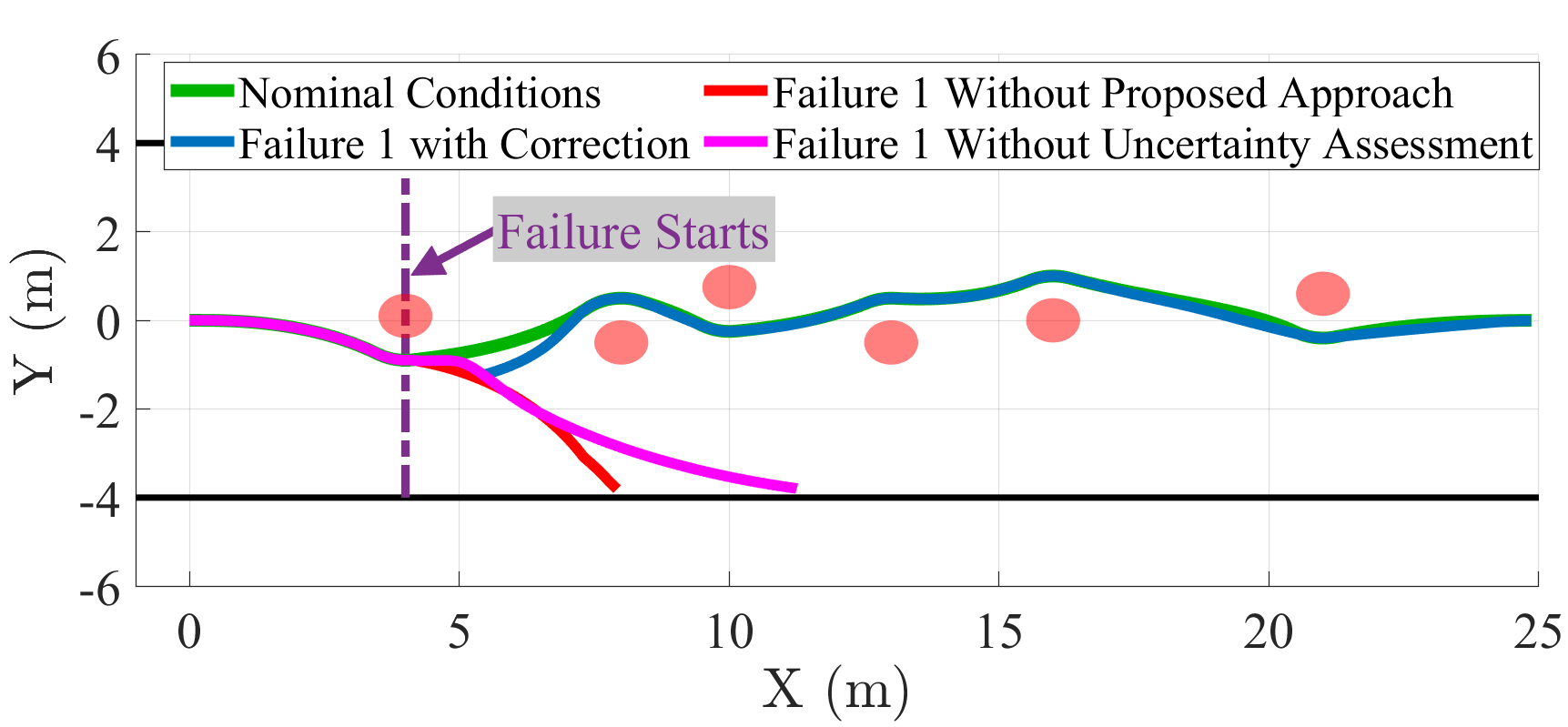}
    \caption{Simulation trajectories.}
    \label{fig:sim1}
    
\end{figure}

In MATLAB simulations, the robot was tasked to move through an environment from an initial point at $(0,0)$ to a goal at $(25,0)$ while avoiding obstacles in the presence of failures. The simulation training set consisted of 5 failures, including: $f_1$: velocity based steering loss where higher speeds lead to more deviations, $f_2$: steering loss of 30\%, $f_3$: icy conditions causing the robot to spin, $f_4$: windy conditions adding bias to the robot position, and $f_5$: steering loss of 10\%. The MPC had a horizon of $N=5$s, and a known map of the environment was used to define obstacle avoidance constraints for the MPC. Shown in Fig.~\ref{fig:sim1} is a comparison between robot behavior under nominal conditions and behaviors under a failure.

The nominal trajectory (green) shows the desired behavior of the robot, while the red line shows that a robot under a failure with no correction collides with a wall. In blue, we show that under our approach, the robot is able to correct its behaviors and converge to nominal behaviors. The magenta path depicts a version of our approach in which the uncertainty assessment and controller validation are removed, and only the initial decision of the DT is utilized to select a controller. This type of decision making leads to a collision, affirming the value of uncertainty assessment and controller validation. 
\begin{figure}[h]
	\centering
	\subfigure[Decision making without uncertainty assessment \label{fig:preddata_wrong}]{\includegraphics[width=0.48\textwidth]{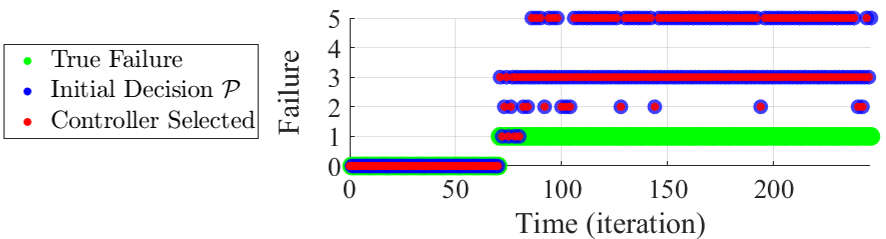}}
	\subfigure[Decision making with uncertainty assessment \label{fig:preddata}]{\includegraphics[width=0.48\textwidth]{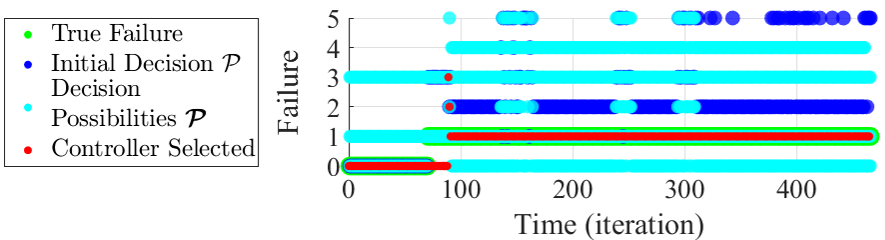}}
	\caption{Comparison of decision making with and without uncertainty assessment and controller validation.}
	\label{fig:simresults}
	\vspace{-15pt}
\end{figure}

\begin{figure}[h]
    \includegraphics[width=0.48\textwidth]{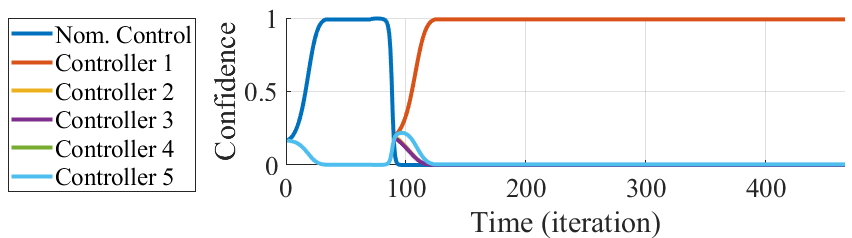}
    \caption{Controller validation results}
    \label{fig:confdata}
    \vspace{-10pt}
\end{figure}

In Fig.~\ref{fig:simresults}, we compare the decision making results of some of the cases listed in Fig.~\ref{fig:sim1}. Fig.~\ref{fig:preddata_wrong} shows the local decisions (blue markers) and controller selections (red markers) without uncertainty assessment and validation. The selected controller always matches the decision $\mathcal{P}$ in this case, and this as seen above, causes a collision as indicated by the magenta line in Fig.~\ref{fig:sim1}. 
In Fig.~\ref{fig:preddata}, the cyan markers show the results of the uncertainty assessment $\bm{\mathcal{P}}$, and despite the poor initial decisions, our approach captures the correct failure. The system tests controllers $c_2$ and $c_3$ when they are safe and learns to instead use $c_1$, which is reinforced positively after deviations below $\eta^*$ are detected, shown by the blue line in Fig.~\ref{fig:sim1}.
In Fig.~\ref{fig:confdata}, we show the confidence in our corrective measures over time, in which system correctly converges to $c_1$. 

\begin{figure}[h]
	\centering
 \subfigure[Trajectory under unknown failure \label{fig:trajdatau}]{\includegraphics[width=0.45\textwidth]{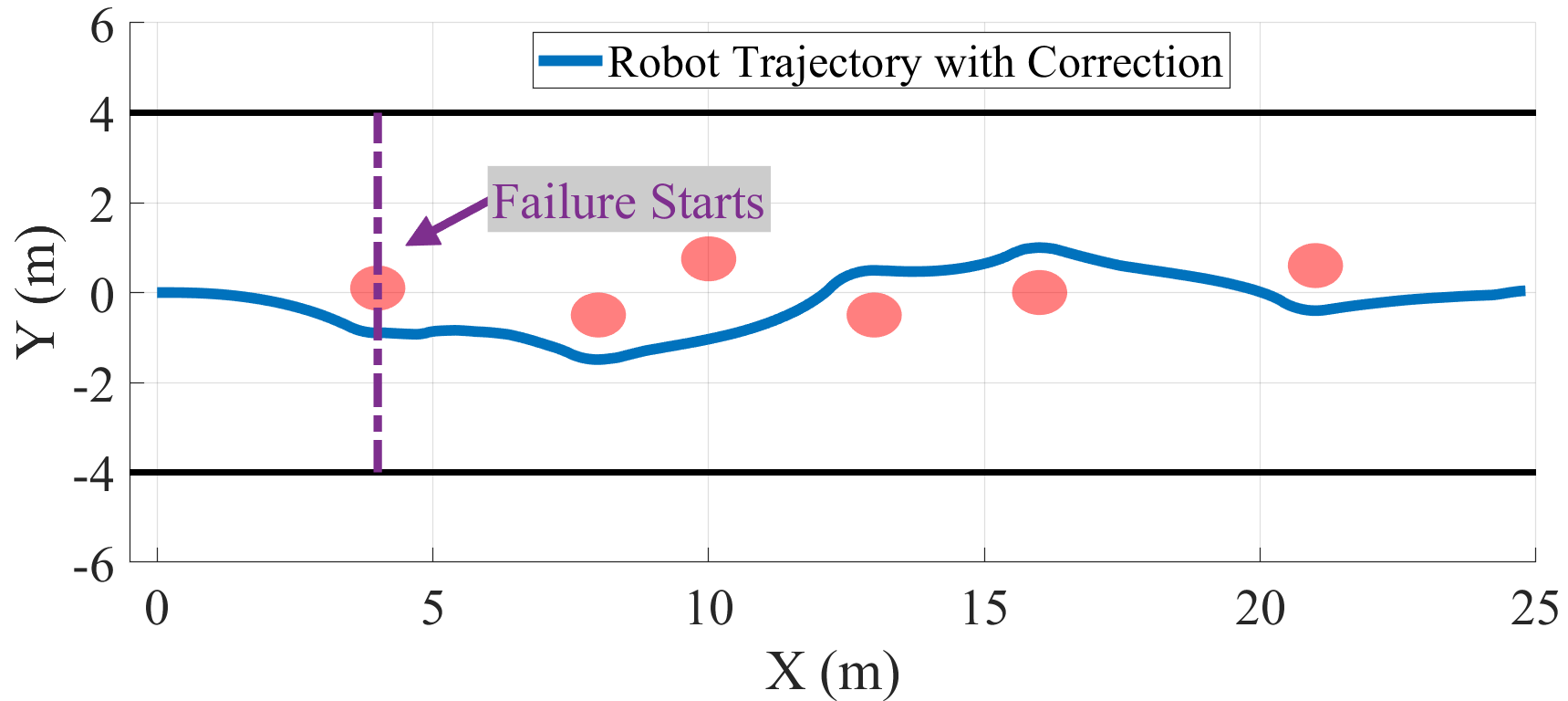}}
	\subfigure[Decision making under unknown failure \label{fig:preddatau}]{\includegraphics[width=0.45\textwidth]{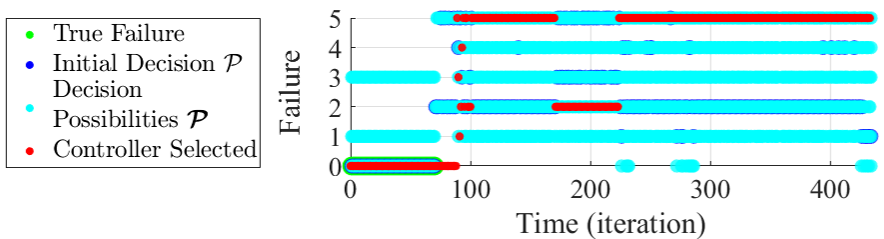}}
	\subfigure[Controller validation results \label{fig:confdatau}]{\includegraphics[width=0.45\textwidth]{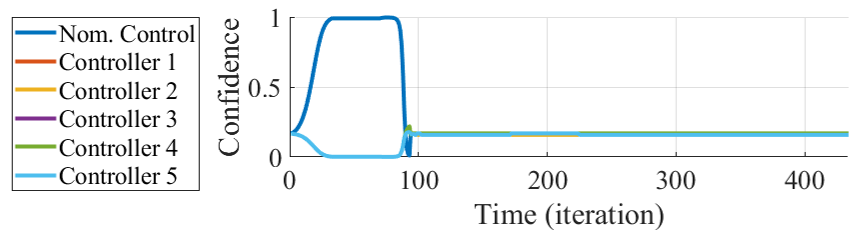}}
	\caption{Results from simulation of unknown failure.}
	\label{fig:unknownsim}
	\vspace{-10pt}
\end{figure}

Another simulation was conducted containing a failure that was not explicitly modeled in training set, but is bounded by failures $f_2$ and $f_5$. Through the results shown in the trajectory in Fig.~\ref{fig:trajdatau} and the data in Fig.~\ref{fig:preddatau}, we observe that the system selects both $c_2$ and $c_5$ at different times since confidence in all controllers is at the initial value (Fig.~\ref{fig:confdatau}) due to repeated negative reinforcement, and the system is using runtime deviations to select a safe controller.

\subsubsection{Hardware Experiments}
\begin{figure}[h]
    \centering
	\subfigure[Snapshots of nominal experiment \label{fig:exp1snap}]{\includegraphics[width=0.43\textwidth,height=2cm]{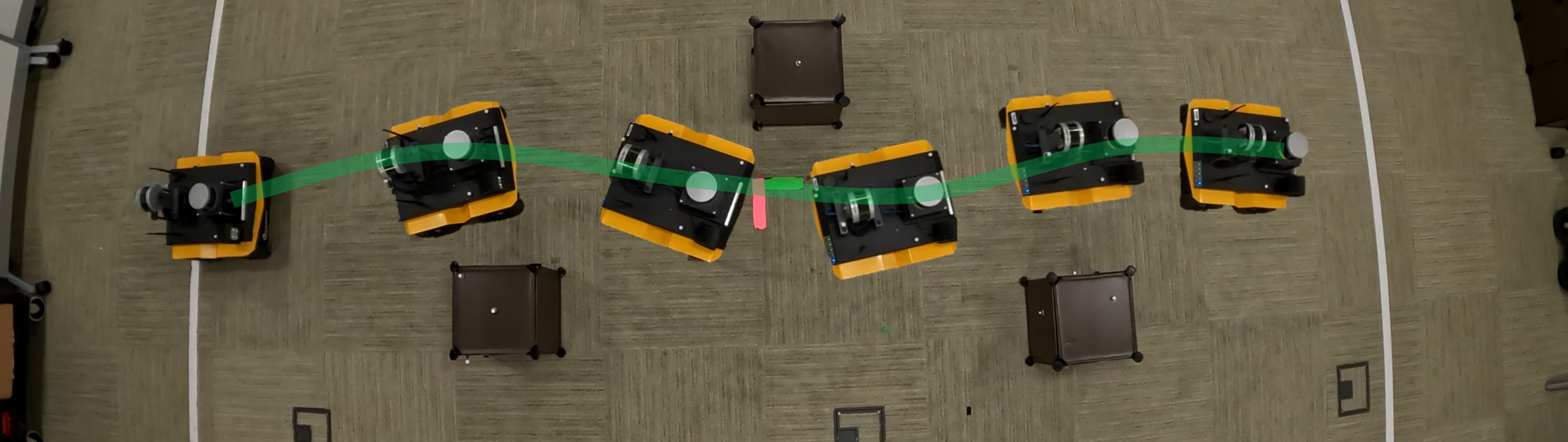}}
 \subfigure[Snapshots of failure case with no recovery \label{fig:exp2snap}]{\includegraphics[width=0.43\textwidth,height=2cm]{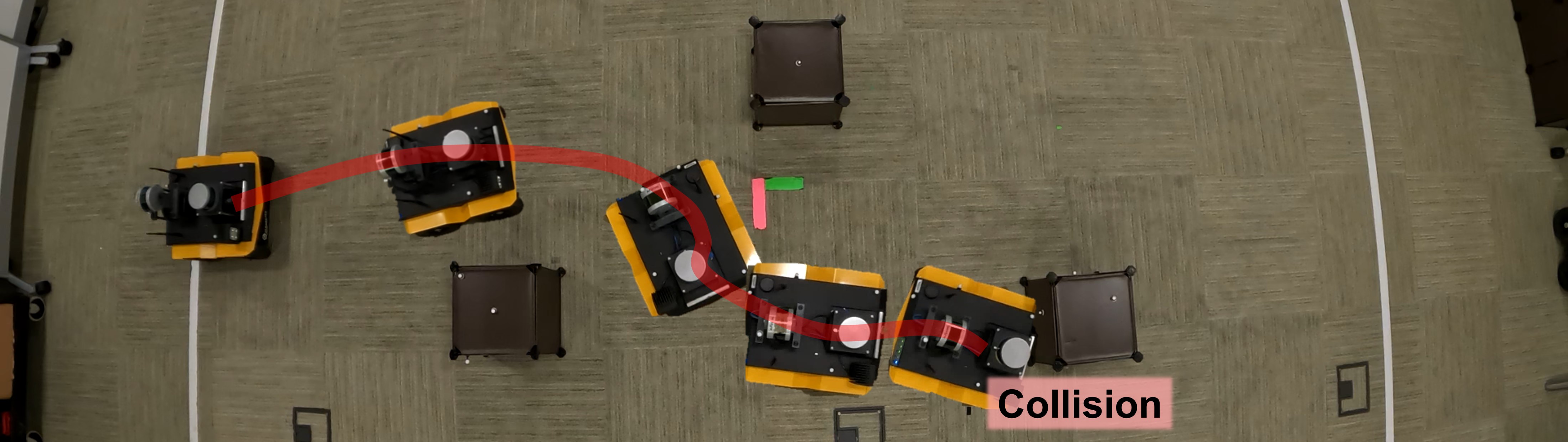}}
 \subfigure[Snapshots of recovery approach \label{fig:exp3snap}]{\includegraphics[width=0.43\textwidth,height=2cm]{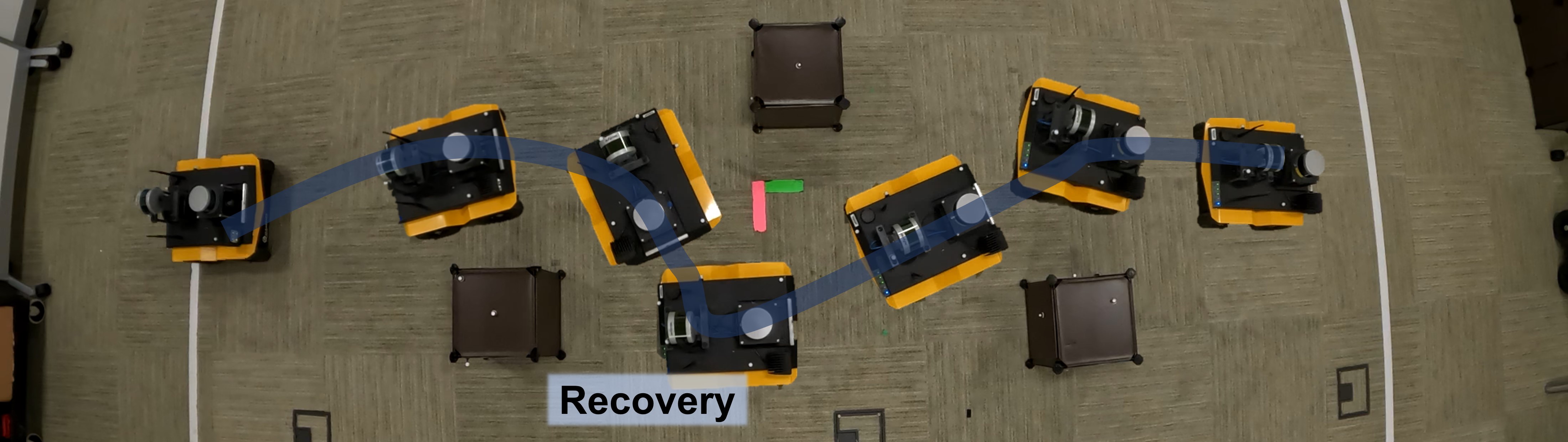}}
	\subfigure[Trajectories of experiments \label{fig:exp1traj}]{\includegraphics[width=0.43\textwidth,height=3cm]{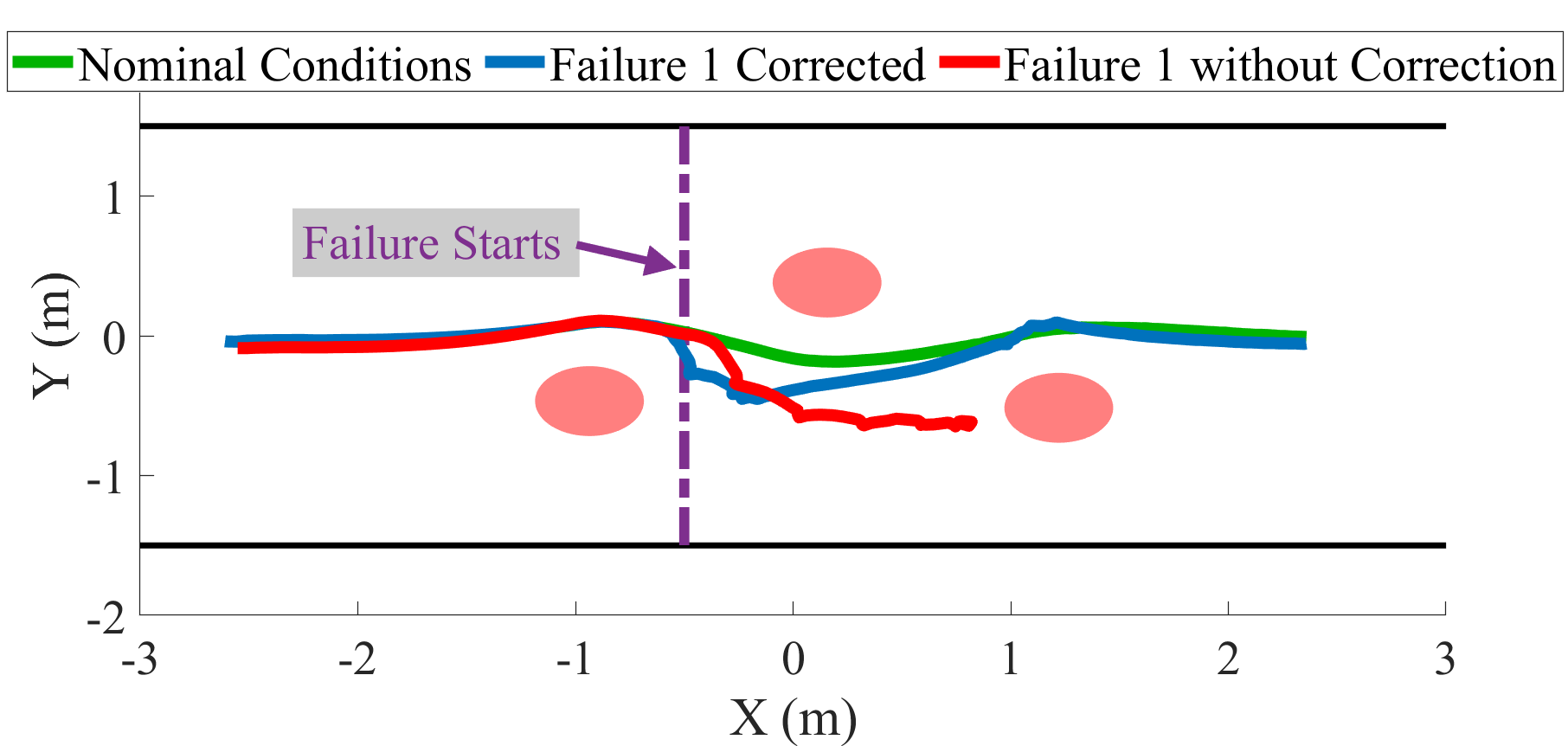}}
    \caption{Snapshots and trajectories for baseline experiments}
    \label{fig:exp1}
    \vspace{-25pt}
\end{figure}

Hardware experiments consisted of 3 failures, which introduce different degrees (10\%, 20\%, and 30\%) of steering loss to the vehicle. Training and testing were done on a Clearpath Jackal UGV in a lab environment with a Vicon Motion Capture system for localization. In these experiments, the failures were injected by artificially adjusting the robot's angular velocity input to reflect steering failures. The MPC horizon was set to $N=3$s, and our approach was executed at $10$hz. The first experiment is similar to the simulation where the robot starting at $(-2.5,0)$ has a task to safely reach a goal at $(2.5,0)$ under a failure that is introduced at runtime. In Fig.~\ref{fig:exp1}, we show that our approach recovers the robot safely, and without our approach, the robot collides with an obstacle.

\begin{figure}[h]
\centering
    \subfigure[Trajectories of experiments \label{fig:exp2traj}]{\includegraphics[width=0.47\textwidth,height=4cm]{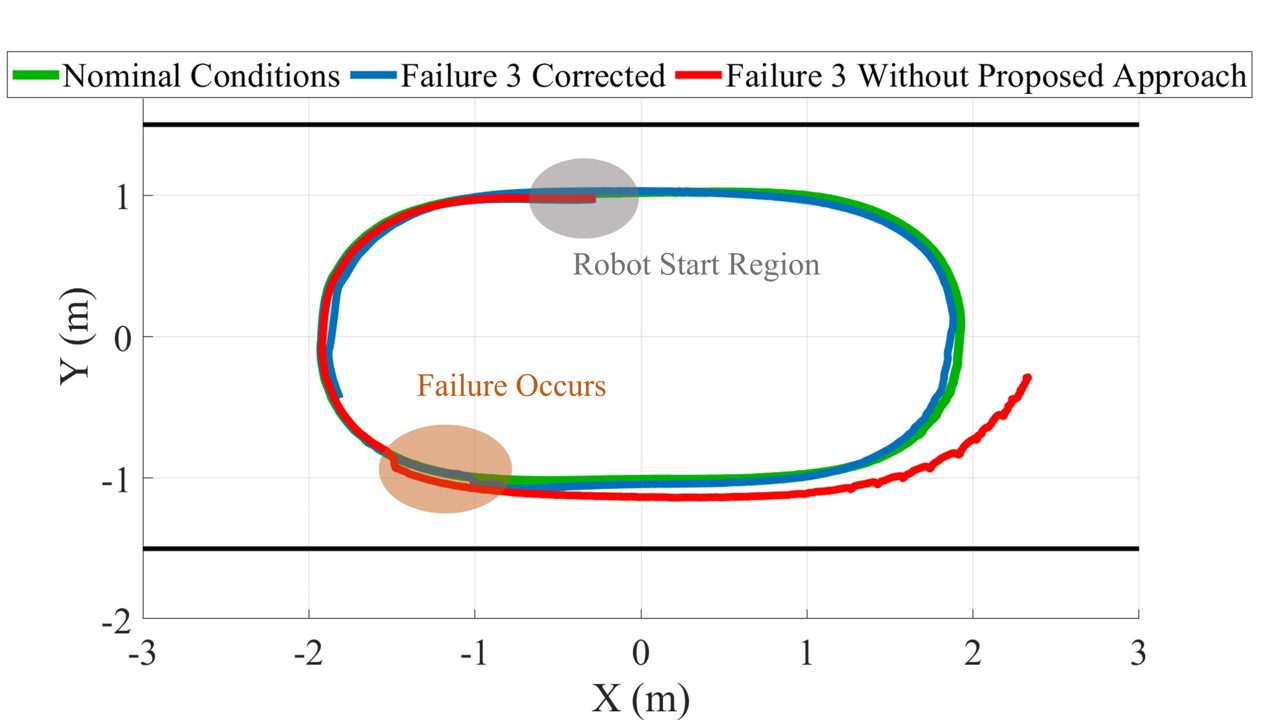}}
    \subfigure[Decision making in ellipse trajectory \label{fig:ellconf}]{\includegraphics[width=0.47\textwidth]{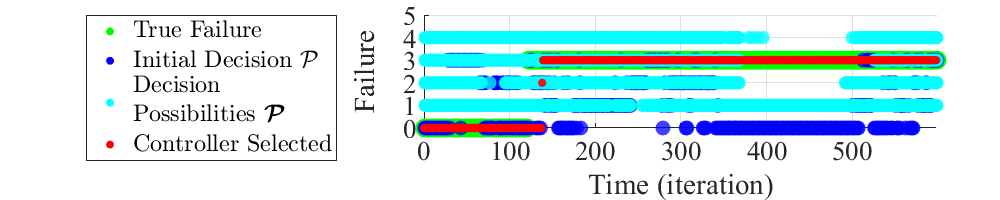}}
    \subfigure[Controller validation results \label{fig:elldata}]{\includegraphics[width=0.47\textwidth]{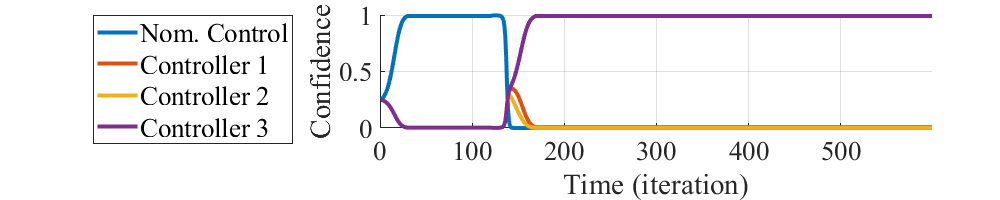}}
	\caption{Results from ellipse experiment.}
	\label{fig:exp2data}
        \vspace{-10pt}
\end{figure}

In the second experiment, the robot is tasked to track an ellipse trajectory to patrol the center of the environment. Trajectories showing the effect of the proposed corrective approach are shown in Fig.~\ref{fig:exp2data}, and confirm that without correction enabled, the robot diverges from its task, while the proposed approach is able to maintain its performance.

\begin{figure}[h]
\centering
\end{figure}

\section{Conclusions} \label{sec:conclusion}

In this work, we have presented a novel approach to handle uncertainties in decision making for recovering an autonomous mobile robot from failures caused by sensor and actuator faults. We design an explainable decision tree-based monitor to detect failures and perturbation based uncertainty assessment to recover the robot to a safe mode of operation. The main benefit of our approach is that it considers and makes use of the uncertainties in the output of a learning component for robot control, promoting safe robot navigation under uncertain and noisy degraded conditions. Dealing with unbounded and unknown failures still remains a challenge in situations where a safe controller may not exist. In future work, we look towards mitigating such issues by using observed deviations to learn the degraded dynamics safely at runtime through system identification methods or reinforcement learning.

\section{Acknowledgement}
This work is based on research sponsored by DARPA under Contract No. FA8750-18-C-0090 and NSF under grant number \#1816591
\bibliographystyle{IEEEtran}
\bibliography{References}

\begin{thebibliography}{10}
\providecommand{\url}[1]{#1}
\csname url@rmstyle\endcsname
\providecommand{\newblock}{\relax}
\providecommand{\bibinfo}[2]{#2}
\providecommand\BIBentrySTDinterwordspacing{\spaceskip=0pt\relax}
\providecommand\BIBentryALTinterwordstretchfactor{4}
\providecommand\BIBentryALTinterwordspacing{\spaceskip=\fontdimen2\font plus
\BIBentryALTinterwordstretchfactor\fontdimen3\font minus
  \fontdimen4\font\relax}
\providecommand\BIBforeignlanguage[2]{{%
\expandafter\ifx\csname l@#1\endcsname\relax
\typeout{** WARNING: IEEEtran.bst: No hyphenation pattern has been}%
\typeout{** loaded for the language `#1'. Using the pattern for}%
\typeout{** the default language instead.}%
\else
\language=\csname l@#1\endcsname
\fi
#2}}

\bibitem{mldecisions}
W.~Schwarting, J.~Alonso-Mora, and D.~Rus, ``Planning and decision-making for
  autonomous vehicles,'' \emph{Annual Review of Control, Robotics, and
  Autonomous Systems}, vol.~1, no.~1, pp. 187--210, 2018.

\bibitem{statest}
P.~Guo, H.~Kim, N.~Virani, J.~Xu, M.~Zhu, and P.~Liu, ``Roboads: Anomaly
  detection against sensor and actuator misbehaviors in mobile robots,'' in
  \emph{2018 48th Annual IEEE/IFIP International Conference on Dependable
  Systems and Networks (DSN)}, 2018, pp. 574--585.

\bibitem{uavdetection}
A.~Keipour, M.~Mousaei, and S.~Scherer, ``Automatic real-time anomaly detection
  for autonomous aerial vehicles,'' in \emph{2019 International Conference on
  Robotics and Automation (ICRA)}, 2019, pp. 5679--5685.

\bibitem{cnnfault}
A.~R. Javed, M.~Usman, S.~U. Rehman, M.~U. Khan, and M.~S. Haghighi, ``Anomaly
  detection in automated vehicles using multistage attention-based cnn,''
  \emph{IEEE Transactions on Intelligent Transportation Systems}, vol.~22,
  no.~7, pp. 4291--4300, 2021.

\bibitem{drlfault}
F.~Huang, \emph{et~al.}, ``A general motion control architecture for an
  autonomous underwater vehicle with actuator faults and unknown disturbances
  through deep reinforcement learning,'' \emph{Ocean Engineering}, vol. 263, p.
  112424, 2022.

\bibitem{esenram}
E.~Yel, \emph{et~al.}, ``Assured runtime monitoring and planning: Toward
  verification of neural networks for safe autonomous operations,'' \emph{IEEE
  Robotics \& Automation Magazine}, vol.~27, no.~2, pp. 102--116, 2020.

\bibitem{incorrectpred}
Q.~M. Rahman, P.~Corke, and F.~Dayoub, ``Run-time monitoring of machine
  learning for robotic perception: {A} survey of emerging trends,''
  \emph{CoRR}, vol. abs/2101.01364, 2021.

\bibitem{drl_uncert}
G.~Kahn, A.~Villaflor, V.~Pong, P.~Abbeel, and S.~Levine, ``Uncertainty-aware
  reinforcement learning for collision avoidance,'' \emph{CoRR}, vol.
  abs/1702.01182, 2017.

\bibitem{uncertaintyquant}
M.~Abdar, \emph{et~al.}, ``A review of uncertainty quantification in deep
  learning: Techniques, applications and challenges,'' \emph{Information
  Fusion}, vol.~76, pp. 243--297, 2021.

\bibitem{samp_slow}
A.~Loquercio, M.~Segu, and D.~Scaramuzza, ``A general framework for uncertainty
  estimation in deep learning,'' \emph{IEEE Robotics and Automation Letters},
  vol.~5, no.~2, pp. 3153--3160, 2020.

\bibitem{bnn1}
A.~Y. Foong, Y.~Li, J.~M. Hern{\'a}ndez-Lobato, and R.~E. Turner,
  ``'in-between'uncertainty in bayesian neural networks,'' \emph{arXiv preprint
  arXiv:1906.11537}, 2019.

\bibitem{bnntut}
L.~V. Jospin, H.~Laga, F.~Boussaid, W.~Buntine, and M.~Bennamoun, ``Hands-on
  bayesian neural networks—a tutorial for deep learning users,'' \emph{IEEE
  Computational Intelligence Magazine}, vol.~17, no.~2, pp. 29--48, 2022.

\bibitem{whyblackbox}
C.~Rudin and J.~Radin, ``Why {Are} {We} {Using} {Black} {Box} {Models} in {AI}
  {When} {We} {Don}\textquoteright{}t {Need} {To}? {A} {Lesson} {From} an
  {Explainable} {AI} {Competition},'' \emph{Harvard Data Science Review},
  vol.~1, no.~2, nov 22 2019, https://hdsr.mitpress.mit.edu/pub/f9kuryi8.

\bibitem{varinfo}
J.~Steinbrener, K.~Posch, and J.~Pilz, ``Measuring the uncertainty of
  predictions in deep neural networks with variational inference,''
  \emph{Sensors}, vol.~20, no.~21, 2020.

\bibitem{activelearn}
A.~T. Taylor, T.~A. Berrueta, and T.~D. Murphey, ``Active learning in robotics:
  A review of control principles,'' \emph{Mechatronics}, vol.~77, p. 102576,
  2021.

\bibitem{jintfault}
B.~Abci, M.~El~Badaoui El~Najjar, V.~Cocquempot, and G.~Dherbomez, ``An
  informational approach for sensor and actuator fault diagnosis for autonomous
  mobile robots,'' \emph{Journal of Intelligent \& Robotic Systems}, vol.~99,
  no.~2, pp. 387--406, 2020.

\bibitem{rnnfault}
J.~Wang, G.~Wu, L.~Wan, Y.~Sun, and D.~Jiang, ``Recurrent neural network
  applied to fault diagnosis of underwater robots,'' in \emph{2009 IEEE
  International Conference on Intelligent Computing and Intelligent Systems},
  vol.~1, 2009, pp. 593--598.

\bibitem{ccdc2019}
X.-Z. Jin, J.-Z. Yu, L.~Zhou, and Y.-Y. Zheng, ``Robust adaptive trajectory
  tracking control of mobile robots with actuator faults,'' in \emph{2019
  Chinese Control And Decision Conference (CCDC)}, 2019, pp. 2691--2695.

\bibitem{mpcsurvey}
S.~Yu, M.~Hirche, Y.~Huang, H.~Chen, and F.~Allgöwer, ``Model predictive
  control for autonomous ground vehicles: a review,'' \emph{Autonomous
  Intelligent Systems}, vol.~1, 12 2021.

\bibitem{rahulral}
R.~Peddi and N.~Bezzo, ``An interpretable monitoring framework for virtual
  physics-based non-interfering robot social planning,'' \emph{IEEE Robotics
  and Automation Letters}, vol.~7, no.~2, pp. 5262--5269, 2022.

\bibitem{catvar}
Y.~S. Kim, ``Comparison of the decision tree, artificial neural network, and
  linear regression methods based on the number and types of independent
  variables and sample size,'' \emph{Expert Systems with Applications},
  vol.~34, no.~2, pp. 1227--1234, 2008.

\bibitem{dtrees}
M.~Krzywinski and N.~Altman, ``\BIBforeignlanguage{English (US)}{Classification
  and regression trees},'' \emph{\BIBforeignlanguage{English (US)}{Nature
  Methods}}, vol.~14, no.~8, pp. 757--758, July 2017.

\bibitem{reach}
J.~Ding, E.~Li, H.~Huang, and C.~J. Tomlin, ``Reachability-based synthesis of
  feedback policies for motion planning under bounded disturbances,'' in
  \emph{2011 IEEE International Conference on Robotics and Automation}, 2011,
  pp. 2160--2165.

\bibitem{bayes}
M.~Castellano-Quero, J.-A. Fern{\'a}ndez-Madrigal, and A.~Garc{\'\i}a-Cerezo,
  ``Improving bayesian inference efficiency for sensory anomaly detection and
  recovery in mobile robots,'' \emph{Expert Systems with Applications}, vol.
  163, p. 113755, 2021.

\end{thebibliography}

\end{document}